# TECHNICAL COMPARATIVE BENCHMARKING STUDY: ADVANCED AI HYBRID METHODS FOR RENEWABLE ENERGY FARM OPTIMIZATION AND FORECASTING

Majid Masoumi
Department of Computer Science
Yazd University
Yazd
Majid.Masoumi@yazd.ac.ir

Asghar Dashtiy
Department of Mechanical Engineering
Yazd University
Yazd
Asghar.Dashtiy@yazd.ac.ir

Mohammad Dehghan
Department of Mechanical Engineering
Yazd University
Mohammad.Dehghan@yazd.ac.ir

Mina Rajabi
Department of Mechanical Engineering
Yazd University
Mina.Rajabi@yazd.ac.ir

## ABSTRACT

Advanced renewable-energy systems present fundamentally different data-driven challenges, ranging from high-dimensional wave energy converter (WEC) layout optimization to noisy, spatially coupled wind-power forecasting. This study provides a comprehensive benchmarking of conventional machine learning (ML), ensemble learning, deep neural networks, recurrent architectures, Transformers, graph-based models, and hybrid ensemble–deep learning approaches under complementary offshore energy scenarios. Three datasets are considered: a large-scale 49WEC dataset containing 63,600 configurations and 149 numerical features, a 16-WEC dataset containing 288,000 layouts across four Australian wave climates, and operational 10-min SCADA measurements from 14 turbines at the Penmanshiel wind farm. For structured WEC-layout data, tree ensembles exhibited a clear advantage over conventional ML and neural predictors because randomized partitioning and boosting efficiently captured nonlinear layout–power interactions without requiring explicit feature representation learning. At the Perth site, Extra Trees (Extra-TD) was the strongest model, achieving considerable results. Relative to the MLP baseline, this corresponds to an approximately 63.7% reduction in MAE, demonstrating the suitability of randomized tree ensembles for high-dimensional structured WEC data. The advantage was taskdependent, however for real multivariate wind-farm forecasting, standalone DNN, LSTM, GRU, and BiLSTM architectures were substantially weaker because temporal representation alone could not fully describe inter-turbine dependencies. STGCN reduced the MAE to approximately 167.0 kW and achieved $R = 0.93$ by explicitly learning spatial and temporal turbine interactions. The best overall forecasting accuracy was obtained by the RF–BiLSTM hybrid, with an MAE of 150.5 kW. Compared with standalone LSTM (MAE = $590.0$ kW), this represents an approximately 75% reduction in MAE, while improving on STGCN by approximately 10.0%. Its superior performance arises from combining variance-reducing nonlinear feature learning by Random Forest with bidirectional temporal representation. In contrast, XGBoost–BiLSTM produced the highest mean correlation but showed substantially greater error variability, demonstrating that maximum correlation does not necessarily imply the most reliable forecast. Finally, the experiments reveal that no single AI architecture is universally optimal: randomized and boosted ensembles are particularly effective for structured WEC surrogate modeling, graph networks become advantageous when explicit spatial interactions dominate, and ensemble–recurrent hybrids provide the strongest balance when nonlinear tabular relationships and temporal dynamics coexist. These findings provide practical guidance for selecting AI architectures according to the physical and statistical structure of offshore renewable-energy problems rather than model complexity alone.

## 1 Introduction

The transition from fossil fuels to low-carbon energy has accelerated considerably over the past decade. Renewable power capacity reached approximately 4,448 GW worldwide by the end of 2024, following the addition of around 585 GW in a single year. Solar and wind accounted for more than 96% of these new renewable capacity additions [1]. Alongside the rapid expansion of onshore renewable generation, increasing attention is being directed toward offshore resources. Offshore wind is already developing at commercial scale in several regions, while wave, tidal, and hybrid offshore systems are attracting growing interest as complementary sources of renewable electricity. The International Energy Agency expects offshore wind deployment to accelerate toward 2030, although financing, supply chain constraints, permitting, and grid integration remain important barriers [2]. Figure 1 highlights substantial regional differences in the 2024 electricity generation mix and in each continent's contribution to global renewable electricity production. Europe and Oceania show relatively high renewable penetration, with renewables accounting for approximately 42.03% and 41.40% of electricity generation, respectively, while the corresponding shares are 36.94% in the Americas, 27.26% in Asia, and 24.74% in Africa. Despite its lower renewable penetration relative to Europe and Oceania, Asia contributes the largest share of global renewable electricity generation (46.66%), reflecting the scale of its overall power system and renewable deployment. The Americas and Europe follow with 26.81% and 17.87%, respectively, whereas Africa and Oceania contribute considerably smaller global shares. These differences demonstrate that renewable penetration and absolute renewable generation are distinct indicators: regions with a high renewable percentage do not necessarily make the largest contribution to global renewable electricity production.

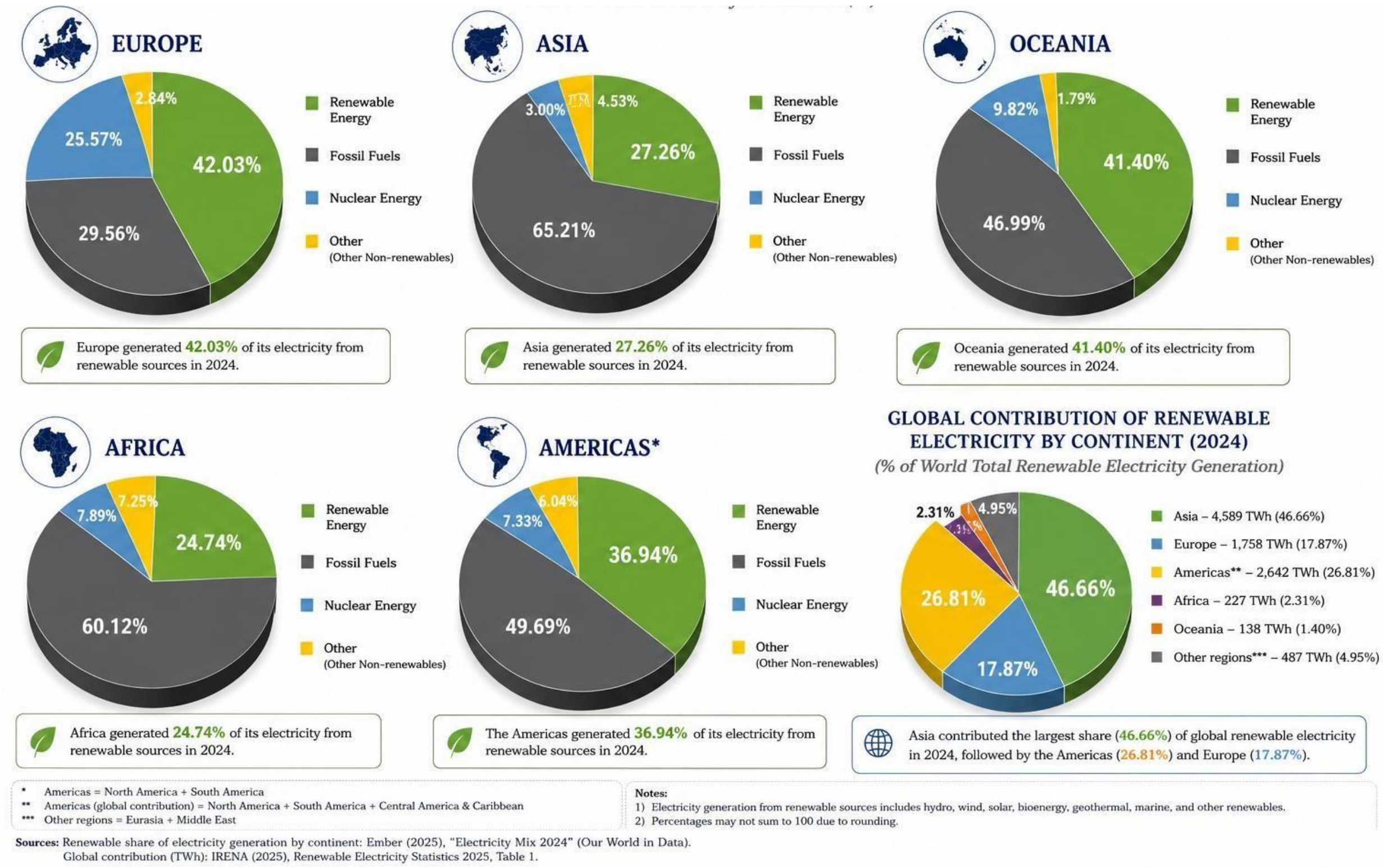

**Figure 1: Renewable electricity generation shares across major continents and their contributions to global renewable electricity generation in 2024.**

Offshore renewable energy (ORE) has several characteristics that make it particularly attractive. Offshore winds are generally stronger and more persistent than those available at many onshore sites, large marine areas provide greater flexibility for utility-scale deployment, and technological progress has enabled increasingly large turbines and offshore structures. Ocean waves also contain substantial energy and can provide a useful complement to wind and solar generation because their temporal characteristics are different. Combining offshore wind, wave energy converters (WECs), storage, and other marine technologies may therefore provide opportunities to improve the use of offshore infrastructure and reduce variability at the system level. These advantages, however, come with considerable technical complexity. Offshore assets operate in environments characterized by changing wind fields, waves, currents, turbulence,
wake effects, hydrodynamic interactions, structural loading, and extreme weather. Installation and maintenance are also expensive [3], which means that relatively small improvements in energy capture, reliability, or operational planning can have meaningful economic consequences. Two computational problems are particularly important in this context: optimization and forecasting.

Optimization arises throughout the life cycle of an offshore energy system. At the planning stage, decisions must be made regarding site selection, farm size, device locations, spacing, electrical infrastructure, and mooring arrangements. At the device and operational levels, turbine control, WEC geometry, power-take-off (PTO) parameters, yaw settings, and maintenance schedules may also require optimization. These variables are strongly coupled. For example, changing the position of one offshore wind turbine affects the wake experienced by downstream turbines, while changing the position of a WEC alters hydrodynamic interactions with neighboring devices. Consequently, maximizing the performance of individual devices does not necessarily maximize the performance of the entire farm. A wide range of optimization techniques has been investigated for these problems. Early studies relied on deterministic search, parametric analysis, genetic algorithms (GAs), and other conventional evolutionary methods. The field subsequently expanded toward differential evolution, covariance matrix adaptation evolution strategy, particle-based methods, local search, cooperative co-evolution, adaptive algorithms, hybrid evolutionary frameworks and multi-objective optimization methods [4]. These approaches are attractive because offshore design problems are often nonlinear, constrained, multimodal, and difficult to solve using gradient information alone. More recently, surrogate models have been incorporated into the optimization loop to reduce the number of computationally expensive hydrodynamic or aeroelastic simulations [5, 6, 7, 8]. Machine learning can approximate the underlying physical simulator, allowing an optimizer to evaluate a much larger number of candidate solutions within a practical computational budget.

**Table 1: Comparative analysis of genetic algorithms and conventional evolutionary optimisation methods for wave energy converter (WEC) farm layout optimisation.**

| Ref. | Year | Study | Optimisation method | Decision variables | Farm size | WEC / hydrodynamic model | Objective / metric | Main benefits | Limitations / weak points |
|---|---|---|---|---|---|---|---|---|---|
| [9] | 2010 | Child and Venugopal, *Optimal configurations of wave ener gy device arrays* | Genetic Algorithm (GA) + Parabolic Intersection (PI) | WEC $x$–$y$ positions and array geometry | Approx. 5 WECs | Generic vertical cylindrical point absorbers; linear potential-flow interaction model | Maximise absorbed power and array interaction | One of the earliest systematic applications of evolutionary optimisation to WEC-array placement; provides comparison between stochastic GA and deterministic PI search. | Mainly small arrays and simplified regular-frequency/directional conditions; computational burden increases rapidly with array dimensionality. |
| [10] | 2011 | Child et al., extension of WECarray and PTO optimisation | Genetic Al gorithm / evolutionary search | WEC positions and PTO-related parameters | Approx. 10 WECs | Point-absorber array with hydrodynamic interaction modelling | Maximise total arra y power | Extended layout optimisation toward simultaneous consideration of device/control parameters and array configuration. | Relatively small farms; simplified environmental assumptions and limited economic and operational constraints. |
| [11] | 2016 | Sarkar et al., *Prediction and optimization of wave energy converter arrays using a machine learning approach* | GA + Monte Carlo + activelearning surrogate | Continuous WEC $x$– $y$ coordinates subject to spatial and bathymetric constraints | 40 WECs | Hydrodynamic interaction model approximated through a statistical / machinelearning emulator | Maximise wave-farm energy performance | Important contribution toward largescale optimisation; surrogate modelling substantially reduces repeated expensive hydrodynamic calculations and allows spatial constraints to be considered. | Optimisation accuracy depends on surrogate training quality; approximate interaction models may fail to capture complex or longrange hydrodynamic effects. |
| [12] | 2017 | Göteman, Wave energy parks with pointabsorbers of different dimensions | Configuration/ parametric optimisation using analytical hydrodynamic model | Device dimensions/ topology and array configuration | 4 WECs | Point absorbers of different dimensions/topologies; analytical multiplescattering model validated against WAMIT | Maximise absorbed power and power-tomass ratio | Demonstrates that heterogeneous WEC dimensions can improve overall park performance; computationally efficient analytical model | Configuration-oriented optimisation rather than a general-purpose global EA; linear potential-flow/heave assumptions. |
| [13] | 2018 | Giassi and Göteman, *Layout design of wave energy parks by a genetic algorithm* | GA wit h continuous an d discrete representations | Continuous or gridbased $x$–$y$ positions; preliminary consideration of WEC geometry and PTO | 4–14 WECs | Point absorbers with analytical multiplescattering formulation | Maximise absorbed power and interaction factor ($q$) | Systematic comparison of continuous and discrete layout representations; provides useful analysis of scalable WEC-park patterns. | Continuous optimisation becomes increasingly expensive as array size grows; wave and hydrodynamic assumptions remain relatively simplified. |
| [14] | 2018 | Sharp and DuPont, *Wave energy converter array optimization: A genetic algorithm approach and minimum separation distance study* | Binary G enetic Algorithm | WEC positions and minimum inter-device separation | 5 WECs | Truncated heavin g cylindrical point absorbers with analytical hydrodynamic model | Maximise power whi l e considering interaction factor and array-related cost | Explicitly investigates minimum separation distance and incorporates economic considerations rather than optimising absorbed power alone. | Small farm size; mainly unidirectional wave conditions and simplified economic modelling restrict generalisation to commercialscale farms. |

| [15] | 2018 | Arbonès et al., multiobjective WEC-array optimisation | Multi-objective Evolutionary Algorithm | WEC layout and selected design/control variables | Approx. 9 WECs | Point-absorber array | Simultaneous optimisation of energy and additional design objectives | Provides Pareto-based analysis and exposes tradeoffs between competing objectives rather than producing only one powermaximising solution. | Multi-objective optimisation significantly increases computational cost; selection among Pareto-optimal designs requires additional decisionmaking criteria. |
|---|---|---|---|---|---|---|---|---|---|
| [16] | 2019 | Lyu, Abdelkhalik and Gauchia, *Optimization of dimensions and layout of an array of wave energy converters* | Genetic Algorithm | WEC $x$–$y$ coordinates, radius and draft | 3, 5 and 7 WECs | Cylindrical WECs under irregular waves with different control strategies | Maximise absorbed power / interaction factor | Strong co-design formulation in which device dimensions and layout are optimised simultaneously; demonstrates considerable improvement in array interaction performance. | Evaluated mainly on small arrays; adding geometry variables substantially increases dimensionality and computational expense. |
| [17] | 2022 | Zeng et al., *Hydrodynamic interactions among wave energy converter array and a hierarchical genetic algorithm for layout optimization* | Hierarchical matrix-coded Genetic Algorithm | Hierarchical optimisation of WEC $x$–$y$ positions | Moderate / large arrays | Truncated cylindrical WECs with semianalytical radiation–diffraction modelling and five degrees of freedom | Maximise extracted wave power | Hierarchical decomposition helps reduce optimisation complexity and supports higher-dimensional layout problems; hydrodynamic representation is more detailed than many earlier GA studies. | Sub-problem decomposition may overlook important interactions between different groups of WECs; hydrodynamics are still predominantly linear/semianalytical. |
| [18] | 2022 | ANN-assisted adaptive GA study for WEC-array layout optimisation | Adaptive Genetic Algorithm + Artificial Neural Network surrogate | WEC $x$–$y$ coordinates; adaptive crossover, mutation and population-search parameters | 3–7 WECs | ANN trained using AQWA-generated hydrodynamic data under measured irregular waves | Maximise total captured wave power | ANN surrogate reduces the number of expensive hydrodynamic simulations, while adaptive GA parameters improve search efficiency relative to conventional GA. | Surrogate generalisation outside the training region may be unreliable; tested farm sizes remain relatively small for commercial deployment. |

Forecasting is equally important once offshore systems become operational. Accurate prediction of wind speed, wave conditions, and generated power supports grid scheduling, reserve management, storage operation, electricity trading, maintenance planning, and system control. Forecast errors can propagate directly into operational and economic decisions. This is particularly challenging offshore because the underlying processes are nonlinear, multivariate, nonstationary, and spatially dependent. Statistical approaches such as autoregressive models and exponential smoothing remain useful benchmarks, but their ability to represent complex nonlinear relationships is limited. Machine-learning methods, including random forests, support vector machines, gradient boosting, XGBoost, and LightGBM, have therefore become common alternatives [19]. Deep-learning models extended this capability by learning temporal representations directly from historical observations. Convolutional neural networks, long shortterm memory networks (LSTMs), gated recurrent units (GRUs) [20], and combinations of these architectures have been applied extensively to renewable-energy forecasting [21, 22, 23, 24]. Attention-based architectures have added another dimension to this development. Transformers can capture long-range temporal relationships without the sequential processing required by conventional recurrent networks. Variants incorporating temporal attention, crossattention, sparse attention, and decomposition mechanisms have consequently been investigated for wind and power forecasting [25, 26, 27, 28]. Nevertheless, greater architectural complexity does not automatically translate into better forecasting. Depending on the amount of training data, prediction horizon, input variables, temporal resolution, and level of noise, comparatively simple ML or recurrent models can remain highly competitive. This makes systematic benchmarking important: the practical question is not simply whether a Transformer can forecast renewable power, but when its additional complexity produces a meaningful and reproducible advantage. Spatial dependence is another aspect that cannot be ignored in offshore farms. Turbines and WECs interact with their neighbors through wakes, waves, environmental fields, and shared operating conditions. Graph neural networks (GNNs) provide a natural way to represent these relationships, with individual assets or measurement locations represented as nodes and their physical or statistical relationships represented as edges. Graph-based forecasting models have already shown promise for learning spatial and temporal dependencies in offshore wind systems [29, 30, 31]. Graph Transformers extend this idea by combining graph representations with attention mechanisms and may be particularly useful when interactions vary with wind direction, sea state, or operating regime.

**Table 2: Comparative analysis of simple evolutionary, heuristic, deterministic and layout-search methods for wave energy converter (WEC) farm layout optimisation.**

| Ref. | Year | Study | Optimisation method | Decision variables | Farm size | WEC / hydrodynamic model | Objective / metric | Main benefits | Limitations / weak points |
|---|---|---|---|---|---|---|---|---|---|
| [32] | 2010 | Babarit, investigation of long-distance hydrodynamic interactions between wave energy converters | Parametric / analytical search | Inter-device separation distance and relative positioning | 2 WECs | Interacting wave energy converters modelled under regular and irregular wave conditions | Quantify hydrodynamic interaction and identify favourable separation distances | Provides fundamental understanding of how interaction effects decay with increasing WEC separation and establishes physical principles subsequently used in array-layout optimisation. | Limited to two-device interaction analysis; does not address highdimensional farm-layout optimisation or large commercial arrays. |
| [33] | 2012 | Borgarino, Babarit and Ferrant, *Impact of wave interaction effects on energy absorption in large arrays of wave energy converters* | Parametric configuration search / prescribedlayout comparison | Array geometry and inter-device spacing | 9, 16 and 25 WECs | Generic heaving cylindrical and surging barge-type WECs with hydrodynamic interaction modelling | Evaluate total energy absorption and array interaction effects | One of the early studies considering relatively large WEC arrays; systematically compares square and triangular configurations and demonstrates the strong effect of spacing on annual energy production. | Optimisation is restricted to predetermined geometric configurations rather than unrestricted continuous WEC positioning. |
| [34] | 2013 | Vicente et al., comparative investigation of wave energy park layouts | Numerical configuration search | Layout geometry, WEC separation and relative arrangement | 12 WECs | WEC array evaluated using multiple numerical hydrodynamic modelling approaches | Maximise constructive hydrodynamic interaction and absorbed power | Comparison of several configurations including line, triangular, square, hexagonal and offset arrangements provides useful physical insight into favourable WEC spacing and orientation. | Search space is restricted to predefined configurations; continuous $x$– $y$ optimisation and broader designvariable interactions are not considered. |
| [35] | 2013 | Engström et al., *Performance of large arrays of point absorbing direct-driven wave energy converters* | Configuration comparison / stochastic layout analysis | Circular, rectangular and random WEC arrangements | 32 WECs | Direct-driven pointabsorbing WEC array with hydrodyn amic interactions | Maximise energy production and investigate aggregate power fluctuations | Considers a relatively large farm and analyses not only total energy capture but also power smoothing and array-level fluctuations, which are relevant to grid integration. | Does not perform comprehensive global optimisation of continuous device locations; conclusions are dependent on a limited set of candidate layouts. |
| [36] | 2014 | de Andrés et al., *Factors that influence array layout on wave energy farms* | Parametric layout optimisation | Array geometry, separation distance, wave direction and number of devices | Small arrays, typically 2–4 WECs | Heaving WECs simulated in the time domain under irregular wave conditions | Assess power absorption and identify favourable farm geometries under different wave climates | Highlights the influence of wave directionality and demonstrates that triangular arrangements can be favourable under multidirectional conditions whereas square layouts may perform well under more directional seas. | Small farm sizes and use of predefined layout families limit scalability and the ability to identify globally optimal unrestricted layouts. |
| [37] | 2015 | Noad and Porter , *Optimisation of arrays of flap-type oscillating wave surge converters* | Multidimensional numerical opt imisation | Inter-device spacing, array configuration and selected device parameters | 3 and 5 WECs | Flap-type oscillating wave surge converters with detailed hydrodynamic modelling | Maximise absorbed wave power and constructive interaction | Provides high-quality devicespecific hydrodynamic optimisation and demonstrates that appropriately selected spacing can significantly improve array-level energy absorption. | Scalability to large farms is limited because of expensive hydrodynamic calculations and rapidly increasing optimisation dimensionality. |
| [38] | 2015–2017 | Moarefdoost et al., WEC farm layout optimisation using mathematical programming | Heuristic initialisation + Sequential / Iterative Quadratic Programming | Continuous WEC $x$– $y$ positions | Approximat 2–15 WECs | elyPoint-absorber array using approxim ate hydrodynamic interaction models | Maximise total absorbed power | Combines heuristic initialisation with computationally efficient gradient-based local optimisation, providing faster convergence than exhaustive or purely | Strong sensitivity to initial solutions and local optima; gradientbased search may perform poorly for highly multimodal layout landscapes and |
| | | | | | | | | populationbased optimisation in some cases. | complex hydrodynamic interactions. |
| [39] | 2016 | Sinha et al., comparative analysis of alternative WECarray arrangements | Numerical / parametric configuration comparison | Array arrang ement and incident-wave direction | 12 WECs | WEC array hydrodynamics evaluated using WAMIT-based modelling | Compare absorbed power under different configurations and incident-wave directions | Demonstrates the strong directional sensitivity of array performance and provides useful evidence for selecting farm orientation relative to dominant wave directions. | Only a limited number of predetermined arrangements are evaluated; no unrestricted evolutionary or continuous optimisation is undertaken. |
| [40] | 2017 | Bozzi et al., *Wave energy farm design in real wave climates: The Italian offshore* | Site-specific parametric optimisation | Array geometry, orientation and interdevice spacing | 4 WECs | Time-domain WECarray model evaluated using realistic Italian offshore wave climates | Maximise annual energy production under realistic environmental conditions | Important site-specific study incorporating realistic directional wave climates and comparing linear, square and rhombus arrangements over multiple device-separation distances. | Small farm size and finite candidatelayout set limit applicability to largescale continuous farm optimisation. |
| [41] | 2017 | Tay and Venugopal, optimisation of oscillating wave surge converter arrays | Modified ev olutionary / numerical parameter search | Device spacing, relative position and array configuration | Up to 12 WECs | Oscillating wave surge converter array under regular, long-crested and short-crested waves | Maximise energy extraction and constructive wave interaction | Shows that optimum array spacing is strongly influenced by wave scattering, transmission and directionality and extends layout analysis beyond conventional point absorbers. | Restricted array geometry and limited optimisation variables reduce generalisability to arbitrary commercial-scale layouts. |
| [42] | 2017 | Wu et al., numerical investigation and optimisation of WECarray parameters | Parametric / numerical optimisation | Inter-device separation, incident-wave parameters and array arrangement | Small to moderate array | Hydrodynamic WECarray model accounting for wavedevice interactions | Maximise absorbed wave energy and analyse sensitivity to environmental parameters | Provides physically interpretable relationships between environmental conditions, separation distance and array performance that can guide optimisation bounds and constraints. | Limited exploration of unrestricted $x$–$y$ placement and no sophisticated strategy for handling highly multimodal search landscapes. |
| [43] | 2018 | López-Ruiz et al., lifecycle-oriented analysis of WEC farm layouts | Configuration and spacing optimisation | Aligned, stagger ed and arrow-type arrang ements together with inter-device spacing | 9 WECs | Wave-energy farm model incorporating energy-production and lifecycle considerations | Improve long-term energy performance and evalua te layoutrelated lifecycle implications | Connects hydrodynamic layout selection with longer-term farm performance; arrow-type layouts can provide favourable energy capture compared with conventional aligned configurations. | Design space remains restricted to several predefined geometries and does not allow completely free device placement. |
| [44] | 2019 | McGuinness and Thomas, constrained optimisation of WEC array configurations | Sequential Quadratic Programming / constrained nonlinear optimisation | Continuous WEC locations subject to motion and separation constraints | 5 WECs | Point-absorber model evaluated over a range of incidentwave directions | Maximise absorbed power while satisfying physic al con- straints | Introduces explicit physical and motion constraints into continuous layout optimisation and provides computationally efficient local refinement. | As a local optimisation method, performance can depend strongly on the initial layout and the approach may converge to suboptimal local solutions in highly multimodal landscapes. |
| [45] | 2019 | Yang et al., integrated assessment of alternative WEC farm arrangements | Comparative optimisation / multi-criteria evaluation | WEC arrangement, farm geometry and spacing | 10 WECs | WEC farm model including waves, current / environmental loading and mooringrelated considerations | Evaluate power production, economic performance and mooring fatigue | Provides a systems-level perspective by considering energy production together with cost and mooringrelated structural performance rather than maximising power alone. | Optimisation is restricted to a small set of candidate configurations and therefore cannot guarantee globally optimal continuous layouts. |

| [46] | 2021 | *Layout optimization of heaving wave energy converter linear arrays in front of a vertical wall* | Constrained numerical optimisation | WEC positions along the wall and interdevice spacing | Multiple WECs | Heaving WEC array with wall–wave– device hydrodynamic interaction under site-specific wave and water-depth conditions | Maximise absorbed wave power by exploiting wallinduced and interdevice interactions | Demonstrates that coastal structures can substantially modify array hydrodynamics and identifies favourable device clustering and positioning close to a reflecting boundary. | Specialised wall-integrated configuration limits generalisation to openocean wave farms; optimisation space is effectively one-dimensional compared with unrestricted offshore layouts. |
|---|---|---|---|---|---|---|---|---|---|

A more recent development is the emergence of pretrained and foundation models for time series. Models such as Chronos, TimesFM, and Lag-Llama have been trained using large collections of time-series data and are designed to transfer knowledge to previously unseen forecasting tasks [39, 40, 41]. This is potentially important for renewable energy, where a conventional model is often retrained separately for every farm, turbine, geographical region, and forecasting horizon. A sufficiently general pretrained model could reduce this dependence on task-specific training and provide useful predictions when historical data are limited. Early studies of foundation models and large pretrained models in the energy domain are encouraging [42, 43], but evidence for offshore applications remains limited. Their robustness under extreme conditions, physical consistency, computational requirements, uncertainty calibration, and ability to generalize between geographically different offshore sites still require careful investigation. The increasing use of artificial intelligence also creates an opportunity to connect forecasting and optimization rather than treating them as independent tasks. Forecasting models can provide environmental and power information to an optimizer, while ML models can serve as computationally efficient surrogates for expensive physical simulations. Physics-informed learning [47] offers another route by embedding physical knowledge into data-driven models rather than relying entirely on observed data [44]. Ultimately, these developments could support digital representations of offshore farms in which forecasting, control, uncertainty estimation, and optimization operate together.

Despite considerable progress in each of these areas, it remains difficult to draw general conclusions from the existing literature. Many studies introduce a new algorithm and demonstrate its performance using one dataset, one farm, or a limited number of baseline methods. Optimization studies may use different computational budgets, population sizes, stopping criteria, physical simulators, and constraints. Forecasting studies similarly differ in preprocessing, traintest partitioning, prediction horizon, error metrics, and hyperparameter tuning. Under these conditions, improvements reported in separate publications cannot be interpreted as direct evidence that one algorithm is generally superior to another. This issue becomes more important as increasingly sophisticated AI architectures are introduced. A model that achieves the lowest RMSE on one offshore wind dataset may require substantially greater training time and memory than a conventional ML model while providing little improvement on another dataset. Similarly, an evolutionary optimizer may obtain an excellent WEC layout for a small array but become impractical when the number of devices and decision variables increases. Therefore, accuracy or final objective value alone provides an incomplete picture. Computational cost, convergence behavior, robustness, scalability, stability across independent runs, uncertainty, and sensitivity to data characteristics should also be considered.

Against this background, this study provides a comprehensive comparative investigation of AI-based forecasting and optimization for offshore renewable-energy systems. Rather than focusing on the introduction of a single new algorithm, the objective is to determine how different methodological families behave under comparable experimental conditions and to identify where their advantages and limitations become important.

The main contributions of this work are:

1. Broad and consistent benchmarking. Statistical methods, conventional ML, deep neural networks, attention/Transformer architectures, graph-based approaches, and emerging advanced models are evaluated using consistent forecasting protocols across multiple offshore renewable-energy datasets.
2. Comprehensive optimization assessment. Representative conventional evolutionary algorithms, advanced evolutionary methods, local-search approaches, hybrid frameworks, and surrogate-assisted strategies are compared under controlled computational conditions for offshore design and layout problems.
3. Evaluation across different problem characteristics. The analysis considers different datasets, prediction horizons, environmental conditions, farm configurations, and optimization landscapes rather than drawing conclusions from a single case study. Comparison beyond predictive accuracy.
4. Forecasting models are examined in terms of accuracy, robustness, computational requirements, stability, and generalization, while optimization algorithms are assessed using solution quality, convergence, computational effort, scalability, and statistical consistency.
5. Critical identification of strengths and weaknesses. The study investigates why particular model families perform well or poorly under different conditions, providing practical guidance for selecting methods rather than simply reporting a global ranking.

The intention is therefore not to identify a universally “best” AI algorithm. Such a conclusion would be unrealistic given the diversity of offshore renewable-energy problems. Instead, this work seeks to establish which methods are most effective under particular data, physical, and computational conditions, what trade-offs accompany their use, and where further methodological development is genuinely needed. This comparative perspective can provide a more reliable basis for the development of accurate, scalable, and practically useful intelligent systems for future offshore renewable-energy applications.

# 2 Offshore Renewable Energy Data

Three datasets were used to provide a balanced evaluation of the forecasting and optimization methods considered in this study. Two datasets represent wave energy farm layout problems, while the third contains operational measurements from a real offshore wind farm. This combination allows the models to be examined under quite different conditions, ranging from high-dimensional WEC layout [48, 49] optimization to multivariate wind-power forecasting using real SCADA data [50] .

## 2.1 Large-scale wave energy farm dataset

The Large-Scale Wave Energy Farm dataset [48] was developed to support data-driven modeling and optimization of large WEC arrays. It contains 63,600 samples and 149 numerical features, with each sample representing one candidate farm configuration. The farms consist of 49 WECs, making this a relatively large optimization problem compared with many earlier WEC studies. In particular, describing the position of every device requires 98 spatial variables ($x$ and $y$ coordinates), in addition to information describing the power generated by individual converters and the overall farm performance.

The main target is the total absorbed power of the farm, while the individual WEC power outputs and interaction information provide additional insight into the hydrodynamic behavior of different layouts. This dataset is particularly useful for testing surrogate models because evaluating large WEC arrays directly through hydrodynamic simulations can be computationally expensive. A sufficiently accurate ML model can learn the relationship between the farm layout and its power output and can then be incorporated into an optimization process to evaluate candidate layouts more efficiently.

## 2.2 Wave energy converter dataset

The second dataset [49] considers a smaller WEC farm but covers a much larger number of candidate layouts. It contains 288,000 samples with 49 continuous features and represents four real wave scenarios along the southern Australian coastline: Sydney, Adelaide, Perth, and Tasmania. Each farm contains 16 fully submerged three-tether CETO WECs.

The main inputs describe the spatial arrangement of the converters through their $x$-$y$ coordinates, together with information related to the power captured by the individual devices. The primary target is again the total absorbed power of the farm. The four wave scenarios provide an additional level of diversity because the same general layout problem can be studied under different environmental conditions.

Using both WEC datasets is useful for evaluating scalability. The 16-WEC dataset provides a large number of observations across several wave environments, whereas the 49-WEC dataset introduces a substantially larger spatial search space. The two datasets therefore allow us to examine how prediction and optimization methods behave as the size and complexity of the wave farm increase. Fig 2 illustrates representative optimized layouts for 16-WEC farms under the Perth and Sydney wave conditions. The spatial distributions demonstrate how WEC placement influences hydrodynamic interactions and overall farm performance, while the color of each WEC represents its individual absorbed power. The variations in total power and $q$-factor further highlight the strong dependence of farm performance on both site-specific wave conditions and array configuration.

## 2.3 Penmanshiel wind farm dataset

The Penmanshiel Wind Farm dataset [50] was used for the wind-energy forecasting experiments. In contrast to the WEC datasets, Penmanshiel contains measurements collected from a real operating wind farm in the United Kingdom. The dataset provides measurements at 10-minute intervals from 2016 to mid-2021 and covers 14 Senvion MM82 wind turbines (with WT03 not included in the released data).

The SCADA measurements provide a detailed description of both environmental conditions and turbine operation. The available variables include wind speed and direction, active power, rotor and generator speeds, nacelle and vane positions, electrical measurements, temperatures, and other turbine operating variables. The main target considered in this study is power generation, with historical multivariate SCADA measurements used to predict future turbine power.

The Penmanshiel dataset is particularly valuable for this comparison because it introduces many of the difficulties encountered in practical forecasting. Unlike simulation-generated data, real SCADA measurements contain noise, missing observations, changing operating regimes, and nonlinear relationships between atmospheric conditions and turbine response. Moreover, measurements from multiple turbines make it possible to investigate not only temporal dependencies but also spatial relationships across the wind farm, which is particularly relevant when evaluating graph-based and spatiotemporal learning models. As shown in Fig. 3, the wind-rose distributions reveal clear differences in the prevailing wind direction and wind-speed frequency across the four study locations. These sitespecific wind characteristics highlight the spatial variability of the available wind resource and provide important environmental context for the subsequent forecasting and optimization analyses. Totally, the three datasets provide complementary test environments. The two WEC datasets focus on learning the relationship between farm layout, hydrodynamic interactions, and captured power, whereas Penmanshiel represents a real-world multivariate timeseries forecasting problem. Their combined use provides a more meaningful basis for comparing conventional ML, deep-learning, Transformer, graph-based, and advanced optimization methods than relying on a single renewableenergy dataset.

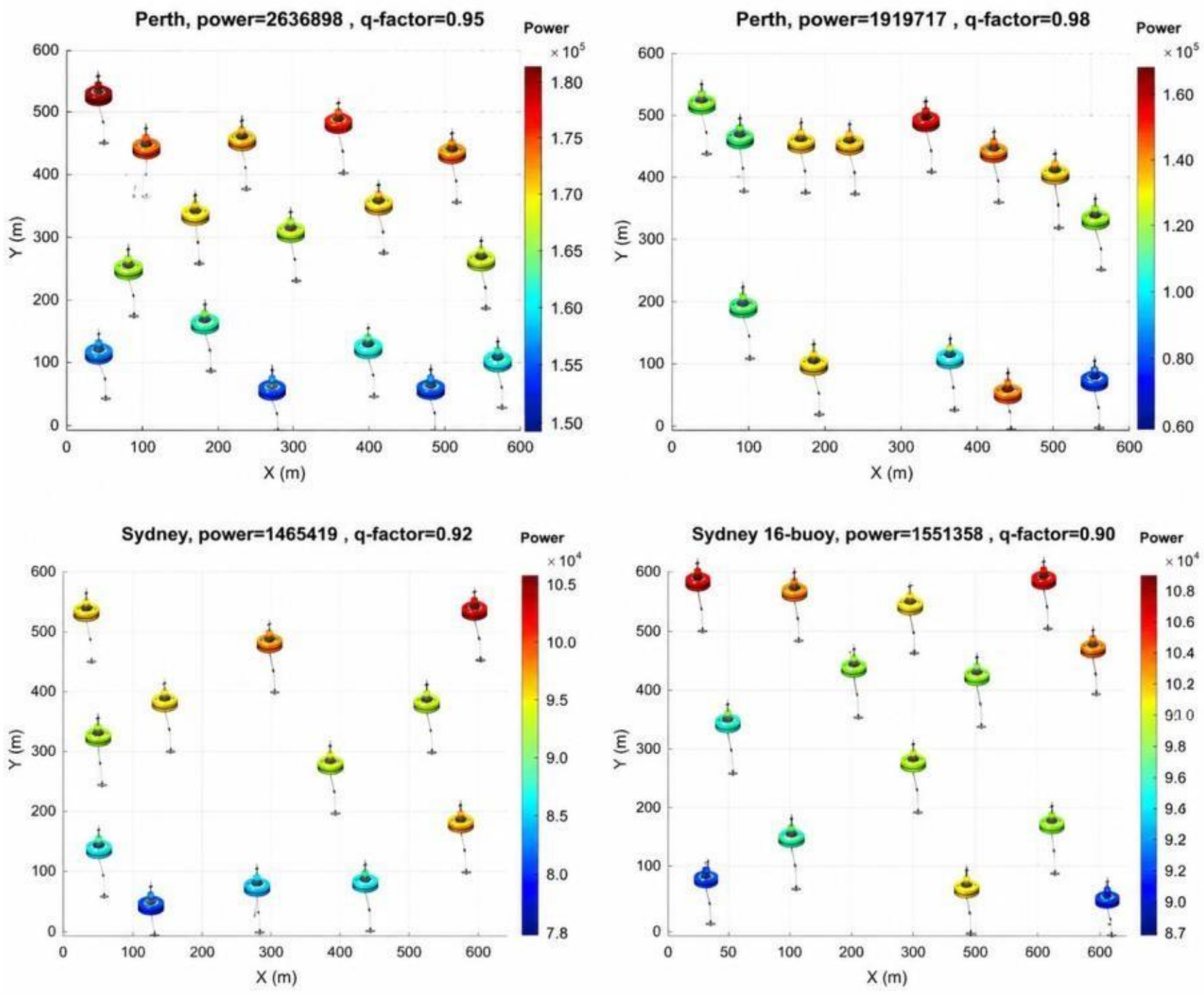

**Figure 2: Optimized layouts of 16-WEC wave farms for Perth and Sydney, where WEC color represents the individual power output and the color bar indicates the corresponding power range.**

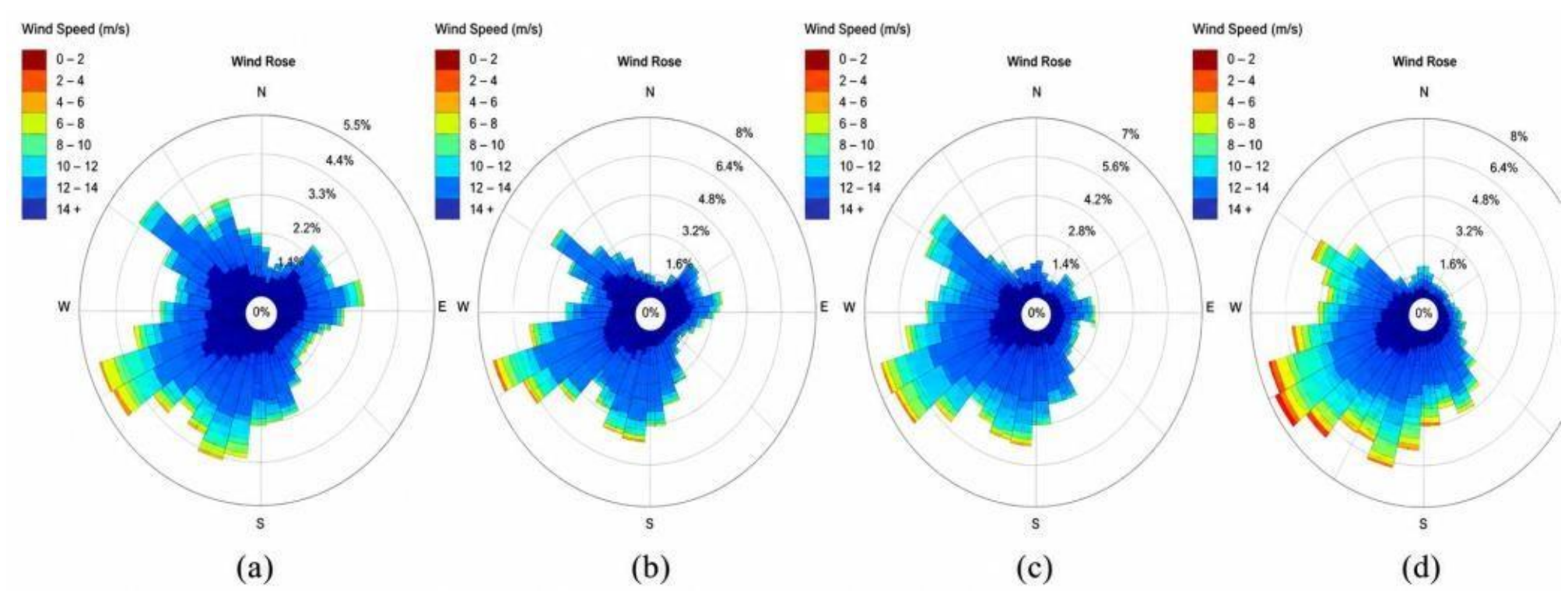

**Figure 3: Wind-rose distributions for the four wave-energy study locations, showing wind direction, occurrence frequency, and wind-speed ranges using the Jet colormap.**

# 3 Advanced AI Forecasting Models and Methodology

To establish a comprehensive benchmark for wave-farm power prediction, a diverse set of regression models was considered, ranging from conventional machine-learning methods to advanced ensemble and cooperative learning approaches. The baseline models include the Multi-Layer Perceptron (MLP), Gradient Boosting Regressor (GBR), Huber-loss Gradient Boosting (GBH), Histogram-Based Gradient Boosting (HGBR), Quantile Gradient Boosting (QGBR), and Decision Tree (DT). More advanced learners comprise AdaBoost (Ada-BT), CatBoost (CAT-B), Extra Trees (ExtraTD), LightGBM, TabNet, and XGBoost (XGB), providing different mechanisms for capturing nonlinear relationships and complex interactions within the WEC layout–control design space. In addition, several ensemble strategies were evaluated, including gradient-based voting (VRGD), heterogeneous voting (VCGD), bagged decision trees (Bag-DT), linear stacked regression (SRGE-Linear), and the cooperative stacked gradient ensemble (SXSGE), which combine complementary predictions from multiple base learners to improve generalization and robustness. Collectively, these models provide a progressively more sophisticated benchmark for assessing whether cooperative and multi-model learning can improve the accuracy and reliability of surrogate prediction for complex 16-WEC farm configurations.

## 3.1 Ensemble Learning Models for Tabular Prediction

Ensemble learning provides an effective framework for modelling nonlinear and heterogeneous relationships in structured tabular data. Rather than relying on a single predictive function, an ensemble combines a collection of base learners, $f_m(\mathbf{x})$, to construct a stronger predictor. For a regression problem, a general ensemble can be expressed as

$$\hat{y} = F(\mathbf{x}) = \sum_{m=1}^{M} w_m f_m(\mathbf{x}), \qquad \sum_{m=1}^{M} w_m = 1, \tag{1}$$

where $\mathbf{x} \in \mathrm{R}^p$ represents the input features, $M$ is the number of learners, and $w_m$ controls the contribution of the $m$th learner. Depending on the ensemble strategy, the individual learners can be trained independently, sequentially, or hierarchically. This diversity is particularly valuable for renewable-energy tabular datasets, where relationships among meteorological, operational, spatial, and device-level variables are often nonlinear and involve complex feature interactions.

### 3.1.1 Bagging-based ensemble learning

Bagging-based ensembles primarily improve generalisation by reducing prediction variance. Given bootstrap samples $\mathrm{D}_1,\ldots,\mathrm{D}_M$ generated from the original training set, independent learners are fitted and their predictions are aggregated as

$$\hat{y}_{\mathrm{bag}}(\mathbf{x}) = \frac{1}{M} \sum_{m=1}^{M} f_m(\mathbf{x}; \mathcal{D}_m). \tag{2}$$

Random Forest (RF) extends this principle by introducing additional randomness through feature subsampling during tree construction, while Extra Trees further increases learner diversity by randomising candidate split thresholds. The variance of an ensemble of correlated learners can be approximated by

$$\mathrm{Var}(\bar{f}) = \rho\sigma^2 + \frac{1-\rho}{M}\sigma^2, \tag{3}$$

where $\sigma^2$ denotes the variance of an individual learner and $\rho$ is the average correlation between learners. Equation (3) shows that increasing diversity among sufficiently accurate base learners can reduce the overall prediction variance. This property is particularly useful for renewable-energy datasets containing noisy measurements and heterogeneous operating conditions.

### 3.1.2 Boosting-based ensemble learning

In contrast to bagging, boosting methods construct learners sequentially, with each new learner attempting to reduce the residual errors of the existing ensemble. Gradient boosting can generally be written as

$$F_m(\mathbf{x}) = F_{m-1}(\mathbf{x}) + \eta h_m(\mathbf{x}), \tag{4}$$

where $h_m(\mathbf{x})$ represents the weak learner introduced at iteration $m$ and $\eta$ is the learning rate. The learner is fitted to approximate the negative gradient of the selected loss function,

$$r_{im} = -\left[\frac{\partial L\left(y_i, F(\mathbf{x}_i)\right)}{\partial F(\mathbf{x}_i)}\right]_{F=F_{m-1}}, \tag{5}$$

allowing the model to progressively focus on prediction errors that have not been adequately captured by previous learners.

Modern gradient-boosting approaches further improve this formulation through regularisation and more efficient tree construction. For example, XGBoost optimises a regularised objective of the form

$$\mathcal{L}^{(t)} = \sum_{i=1}^{N} l\left(y_i, \hat{y}_i^{(t-1)} + f_t(\mathbf{x}_i)\right) + \Omega(f_t), \tag{6}$$

where $l(\cdot)$ is the prediction loss and $\Omega(f_t)$ penalises model complexity. LightGBM improves computational efficiency through histogram-based learning and leaf-wise tree growth, whereas CatBoost introduces specialised boosting mechanisms that are particularly useful when heterogeneous or categorical predictors are present. AdaBoost provides another sequential ensemble strategy by progressively increasing attention to observations that are difficult for previous weak learners to predict.

### 3.1.3 Why ensemble models are effective for tabular data

The strong performance of ensemble methods on tabular datasets can be attributed to their ability to naturally represent nonlinear relationships, thresholds, discontinuities, and high-order feature interactions. Tree-based learners recursively partition the input space and can therefore identify different operating regimes without requiring the functional relationship between the predictors and target to be specified beforehand. This property is well suited to renewableenergy data, where power generation can change nonlinearly with wind speed, wave conditions, control states, turbine characteristics, and other environmental or operational variables.

Tree ensembles also generally require less feature scaling and architectural design than conventional neural networks. In a neural model, structured predictors are transformed through successive nonlinear mappings,

$$\mathbf{h}^{(l)} = \phi\left(\mathbf{W}^{(l)}\mathbf{h}^{(l-1)} + \mathbf{b}^{(l)}\right), \tag{7}$$

where $\mathbf{W}^{(l)}$ and $\mathbf{b}^{(l)}$ denote the weights and biases of layer $l$, respectively, and $\phi(\cdot)$ represents the nonlinear activation function. The network must therefore learn an appropriate internal representation of the structured predictors during optimisation. In contrast, tree ensembles directly search for informative feature thresholds and their interactions. For finite-sized structured datasets, this inductive bias can provide an effective balance between model flexibility and generalisation.

Another important advantage is the bias–variance trade-off. Bagging methods mainly reduce variance by averaging diverse learners, whereas boosting methods progressively reduce residual errors and model bias. Modern implementations such as RF, Extra Trees, XGBoost, LightGBM, and CatBoost additionally incorporate feature subsampling, shrinkage, regularisation, or randomisation. These mechanisms can improve robustness when tabular datasets contain noise, outliers, unevenly represented operating regimes, or complex interactions among input variables.

### 3.1.4 Hybrid ensemble–deep learning models

The comparative results further motivate the integration of ensemble learning with sequential deep-learning architectures.

A generic hybrid ensemble–temporal model can be represented as

$$\mathbf{z}_t = f_{\text{ens}}(\mathbf{x}_t), \tag{8}$$

$$\mathbf{h}_t = f_{\text{seq}}\left(\mathbf{z}_t, \mathbf{h}_{t-1}\right), \tag{9}$$

$$\hat{y}_{t+\Delta} = g(\mathbf{h}_t), \tag{10}$$

where $f_{\text{ens}}(\cdot)$ captures nonlinear relationships among the structured predictors, $f_{\text{seq}}(\cdot)$ represents a sequential model such as LSTM, BiLSTM, or GRU, and $g(\cdot)$ maps the learned representation to the future power prediction at forecasting horizon $\Delta$.

This complementary learning mechanism provides a possible explanation for the competitive performance of hybrid approaches such as RF–BiLSTM, CatBoost–BiLSTM, HGBR–BiLSTM, AdaBoost–GRU, CatBoost–GRU, AdaBoost–LSTM, XGBoost–BiLSTM, and XGBoost–LSTM in the present experiments. The ensemble component is effective at extracting nonlinear relationships and interactions from structured predictors, whereas the recurrent component captures temporal dependencies in the sequential measurements. In particular, RF–BiLSTM achieved the strongest overall error performance among the evaluated forecasting models, indicating that nonlinear tabular feature learning and temporal representation can provide complementary information.

Nevertheless, the superior performance of ensemble models should not be interpreted as a universal property of tabular learning. Their effectiveness depends on the sample size, feature structure, temporal organisation, noise level, preprocessing strategy, and hyperparameter configuration. When explicit spatial connectivity or long-range temporal dependencies dominate the prediction problem, graph neural networks and Transformer-based architectures may offer additional advantages. The strong performance of STGCN observed in the present experiments supports this interpretation, since its graph-based architecture explicitly represents inter-turbine spatial dependencies that conventional tree ensembles cannot directly encode. Overall, the comparative results suggest that ensemble learning provides a strong foundation for structured renewable-energy prediction, while hybrid and spatio-temporal architectures become particularly valuable when nonlinear tabular relationships must be combined with temporal and spatial information. Table 3 and 3 shows the compared methods with the hyper-parameters.

## 4 Experimental Results and Comparative Analysis

To establish a comprehensive and methodologically representative comparative framework, the candidate models evaluated in this study were selected and extended based on two recently published studies in wind and wave renewable energy systems [4, 47]. These studies provide relevant benchmarks covering different families of datadriven methods for renewable-energy prediction and forecasting. Building on these works, the present study brings the selected candidates into a unified experimental framework, including conventional machine-learning models, tree-based ensemble methods, deep and recurrent neural networks, Transformer-based architectures, graph-based learning, and hybrid ensemble–deep learning approaches. This selection strategy was adopted to provide methodological diversity rather than favoring a particular algorithmic family, enabling a systematic assessment of how different learning mechanisms respond to structured WEC data and spatio-temporal wind-farm observations. All candidate methods were subsequently implemented and evaluated under consistent data-processing, training, and performance-assessment protocols to improve the fairness and reproducibility of the comparative analysis.

As can be seen in Figure 4, the MAE plot shows clear differences in prediction error among the evaluated models. The advanced ensemble methods generally outperform the conventional approaches, with Extra-TD achieving the lowest MAE, indicating particularly accurate prediction of the 16-WEC farm power output. The $R^2$ plot evaluates how effectively each model explains the variability in total farm power. Higher values indicate stronger predictive capability, with the leading ensemble models approaching unity and therefore capturing most of the variation in the target power. The CCO plot provides an additional measure of agreement between predicted and actual power values. The best-performing models achieve CCO values close to 1, demonstrating strong correspondence between predictions and the reference power data. The TDA plot highlights differences in prediction deviation and reliability, where lower values indicate better performance. Extra-TD records the lowest TDA among the compared models, followed by MLP among the conventional methods, demonstrating that predictive accuracy according to standard error metrics does not necessarily translate directly into the same ranking for deviation-based reliability.

The three two-dimensional performance Figure 5 provide a complementary view of model accuracy, goodness of fit, agreement, and prediction reliability. In the MAE–$R^2$ space, the strongest models are concentrated toward the upperleft region, representing simultaneously low prediction error and high explained variance. A similar pattern is observed for MAE–CCO, where models approaching the upper-left corner provide both lower absolute errors and stronger agreement between predicted and actual farm power. In contrast, the desirable region of the MAE–TDA map is the lower-left corner because both metrics are minimized. Overall, the plots reveal the stronger trade-off profile of advanced ensemble approaches, particularly Extra-TD, which maintains low MAE while achieving favorable $R^2$, CCO, and TDA values. The dispersion of the remaining models across these spaces also demonstrates that ranking models using a single accuracy metric can be misleading, supporting the use of multiple complementary criteria when selecting reliable surrogate models for WEC-farm power prediction.

**Table 3: Summary of the machine-learning, ensemble-learning, and cooperative surrogate models evaluated in this study, together with their principal hyperparameter settings. Parameters reported as "tuned" were selected during model calibration, while the proposed cooperative framework applies multi-objective covariance-adaptation-based hyperparameter optimization.**

| No. | Abbrev. | Method | Main hyperparameters / configuration | Model family |
|---|---|---|---|---|
| 1 | MLP | Multi-Layer Perceptron | 2–3 hidden layers; ReLU activation; Adam optimizer; learning rate tuned | Neural network |
| 2 | GBR | Gradient Boosting Regressor | $n_{\text{estimators}}$ = $100$; learning rate = $0.1$; maximum tree depth =3 | Boosted trees |
| 3 | GB-H | Gradient Boosting with Huber Loss | Huber loss; $n_{\text{estimators}}$ =$100$; learning rate = $0.1$ | Robust boosted trees |
| 4 | HGBR | Histogram-Based Gradient Boosting Regressor | Maximum bins =$255$; early stopping enabled | Histogram-based boosting |
| 5 | Q-GBR | Quantile Gradient Boosting Regressor | Quantile loss with $\alpha$ =$0.5$; $n_{\text{estimators}}$ =$100$ | Quantile boosting |
| 6 | DT | Decision Tree Regressor | CART formulation; maximum depth tuned; minimum samples per leaf tuned | Single decision tree |
| 7 | AdaBoost | AdaBoost Regressor with Decision Trees | $n_{\text{estimators}}$ =$100$; learning rate =$0.1$; decisiontree weak learners | Adaptive boosting |
| 8 | Cat-Boost | CatBoost Regressor | Tree depth =6; learning rate =$0.1$; optimization loss = RMSE | Ordered gradient boosting |
| 9 | ExtraTree | Extra Trees Regressor | $n_{\text{estimators}}$ =$200$; randomized feature/split selection; maximum depth tuned | Randomized tree ensemble |
| 10 | Light-GBM | Light Gradient Boosting Machine | $n_{\text{estimators}}$ =$200$; learning rate =$0.05$; number of leaves tuned | Leaf-wise gradient boosting |
| 11 | Tab-Net | TabNet | Sequential feature-attention mechanism; decision steps =5; Adam optimizer | Deep tabular learning |
| 12 | XGB | Extreme Gradient Boosting | $n_{\text{estimators}}$= $200$; maximum depth = 6; subsample ratio =$0.8$ | Gradient-boosted trees |
| 13 | Voting-RGD | Voting Regressor (GBRbased) | Soft voting across gradient-based regressors; predictions combined at output level | Voting ensemble |
| 14 | VC-GD | Voting Regressor (Cat-Boost + Gradient Boost-ing) | Soft voting ensemble combining CatBoost and gradient-boosting predictors | Heterogeneous voting ensemble |
| 15 | Bag-DT | Bagged Decision Trees | Bootstrap aggregation; $n_{\text{estimators}}$ =$200$; decision trees used as base learners | Bagging ensemble |
| 16 | Stacked-RGE | Stacked Regression with Linear Meta-Learner | Predictions of multiple base learners used as meta-features; linear regression used as the final meta-learner | Stacking ensemble |

The four violin plots (Figure 6) reveal a clear performance gap between the conventional deep-learning models and approaches that explicitly exploit spatial structure or combine complementary learners. The standalone DNN, LSTM, BiLSTM, and GRU models produce relatively high MAE and MSLE and very low EVS, despite the recurrent models achieving moderate-to-high $R$-values. This suggests that learning temporal dependencies alone is insufficient to represent the complex spatial coupling among turbines. In contrast, STGCN substantially improves all major criteria, achieving an MAE of approximately 167 kW, $R$ = $0.936$, and EVS = 0.876. Its stronger performance can be attributed to the explicit graph representation of inter-turbine relationships, which allows spatial dependencies to be learned jointly with temporal dynamics rather than treating turbine observations as independent sequences. JST-Transformer improves upon the conventional recurrent models, particularly in MSLE, but its wider distributions indicate greater sensitivity and lower stability across repeated runs.

The hybrid models provide further evidence that combining complementary learning mechanisms can improve forecasting performance. RF-BiLSTM shows the strongest overall error performance, with the lowest mean MAE (150.5 kW) and the highest EVS (0.882), while maintaining a high $R$-value of 0.939. The randomized ensemble component can capture nonlinear feature interactions and reduce variance, while BiLSTM models sequential dependencies in both temporal directions, providing complementary representations of the SCADA data. CatBoostBiLSTM and HGBRBiLSTM also achieve $R$-values around 0.94, although their larger MAE distributions indicate that strong correlation does not necessarily imply minimum forecasting error. Similarly, XGBoost-BiLSTM obtains the highest mean $R$-value among the evaluated methods but exhibits considerably greater variability in MAE and EVS, suggesting less consistent generalization across repetitions. Overall, the results indicate that spatio-temporal and hybrid ensemble–recurrent architectures are more effective than standalone neural models, because they can simultaneously capture nonlinear feature interactions, temporal dynamics, and, in the case of STGCN, explicit spatial dependencies. The differences across MAE, $R$, MSLE, and EVS also reinforce the importance of evaluating forecasting models using multiple complementary metrics rather than selecting a model from correlation or prediction error alone.

**Table 4: Architecture and hyperparameter configurations of the deep learning, spatio-temporal, and hybrid ensemble models used for wind power forecasting.**

| Model | Model category | Main architecture and hyperparameters |
|---|---|---|
| DNN | Feed-forward DL | Hidden layers = [256, 128, 64]; activation = ReLU; dropout = 0.20; optimizer = Adam; learning rate = $10^{-3}$; batch size = 128; maximum epochs = 200; early stopping = 20 epochs. |
| LSTM | Recurrent DL | LSTM units = [128, 64]; dropout = 0.20; dense layer = 32; optimizer = Adam; learning rate = $10^{-3}$; batch size = 128; maximum epochs = 200; early stopping = 20 epochs. |
| BiLSTM | Bidirectional recurrent DL | BiLSTM units = [128, 64]; dropout = 0.20; dense layer = 32; optimizer = Adam; learning rate = $10^{-3}$; batch size = 128; maximum epochs = 200; early stopping = 20 epochs. |
| GRU | Recurrent DL | GRU units = [128, 64]; dropout = 0.20; dense layer = 32; optimizer = Adam; learning rate = $10^{-3}$; batch size = 128; maximum epochs = 200; early stopping = 20 epochs. |
| STGCN | Spatio-temporal GNN | Graph convolution channels = [64, 64]; temporal kernel size = 3; graph neighbourhood $k$ = 3; dropout = 0.20; optimizer = Adam; learning rate = $10^{-3}$; batch size = 64; maximum epochs = 200. |
| JST-Transformer | Transformer | $d_{\text{model}} = 128$; attention heads = 8; Transformer blocks = 3; feed-forward dimension = 256; dropout = 0.10; optimizer = AdamW; learning rate = $10^{-4}$; batch size = 64; maximum epochs = 200. |
| CatBoost-BiLSTM | Hybrid ensemble–DL | CatBoost: trees = 500, depth = 8, learning rate = 0.05; BiLSTM units = [128, 64]; dropout = 0.20; Adam ($10^{-3}$); batch size = 128; maximum epochs = 200. |
| HGBR-BiLSTM | Hybrid ensemble–DL | HGBR: max iterations = 300, learning rate = 0.05, max leaf nodes = 31, L2 regularisation = 0.1; BiLSTM units = [128, 64]; dropout = 0.20; Adam ($10^{-3}$); batch size = 128. |
| AdaBoost-GRU | Hybrid ensemble–DL | AdaBoost: estimators = 300, learning rate = 0.05; GRU units = [128, 64]; dropout = 0.20; Adam ($10^{-3}$); batch size = 128; maximum epochs = 200. |
| CatBoost-GRU | Hybrid ensemble–DL | CatBoost: trees = 500, depth = 8, learning rate = 0.05; GRU units = [128, 64]; dropout = 0.20; Adam ($10^{-3}$); batch size = 128; maximum epochs = 200. |
| RF-BiLSTM | Hybrid ensemble–DL | Random Forest: trees = 500, max features = sqrt, minimum samples leaf = 1; BiLSTM units = [128, 64]; dropout = 0.20; Adam ($10^{-3}$); batch size = 128; maximum epochs = 200. |
| AdaBoost-LSTM | Hybrid ensemble–DL | AdaBoost: estimators = 300, learning rate = 0.05; LSTM units = [128, 64]; dropout = 0.20; Adam ($10^{-3}$); batch size = 128; maximum epochs = 200. |
| XGBoost-BiLSTM | Hybrid ensemble–DL | XGBoost: estimators = 500, max depth = 8, learning rate = 0.05, subsample = 0.8, column subsample = 0.8; BiLSTM units = [128, 64]; dropout = 0.20; Adam ($10^{-3}$). |
| XGBoost-LSTM | Hybrid ensemble–DL | XGBoost: estimators = 500, max depth = 8, learning rate = 0.05, subsample = 0.8, column subsample = 0.8; LSTM units = [128, 64]; dropout = 0.20; Adam ($10^{-3}$). |

## 5 Conclusions and Future Research Directions

This study presented a comprehensive comparative assessment of artificial intelligence methods for two important offshore renewable-energy tasks: surrogate prediction of wave energy converter (WEC) farm power and short-term wind-power forecasting. The benchmark covered conventional machine learning, tree-based ensemble learning, deep and recurrent neural networks, Transformer-based architectures, graph learning, and hybrid ensemble–deep learning models. Three complementary datasets were considered, including 63,600 configurations of a large-scale 49-WEC farm, 288,000 16-WEC layouts across four Australian wave climates, and real 10-min SCADA measurements from 14 turbines at the Penmanshiel wind farm. This broad experimental setting allowed the models to be examined under substantially different data structures, dimensionalities, and physical operating conditions.

The results demonstrate that model effectiveness is strongly dependent on the structure of the underlying problem. For structured WEC data, tree-based ensemble methods generally provided the strongest predictive performance. At the Perth site, Extra Trees (Extra-TD) achieved an MAE of $4.7 \times 10^4$, $R^2 = 0.96$. Compared with MLP, which obtained an MAE of $1.4 \times 10^5$, Extra-TD reduced the prediction error by approximately 64%. This substantial improvement indicates that randomized tree ensembles are particularly effective at representing the nonlinear thresholds, highorder feature interactions, and heterogeneous relationships present in WEC layout and control variables. Their strong performance also suggests that increasing neural-network complexity does not necessarily improve prediction when the underlying information is predominantly structured and tabular. A different behavior was observed for short-term wind-power forecasting, where both spatial and temporal dependencies become important. The STGCN achieved an MAE of 167 kW, substantially outperforming standalone DNN and recurrent architectures. Its performance highlights the benefit of explicitly representing inter-turbine relationships through a graph while simultaneously learning temporal dynamics. Nevertheless, the strongest overall error performance was obtained by the hybrid RF–BiLSTM model, which achieved an MAE of 150.5 kW, $R = 0.94$. This corresponds to an approximately 74% reduction in MAE compared with standalone LSTM and an approximately 10.0% improvement over STGCN. The result suggests that combining the nonlinear feature-learning and variance-reduction capability of Random Forest with the temporal representation of BiLSTM can provide complementary information that neither component captures as effectively in isolation.

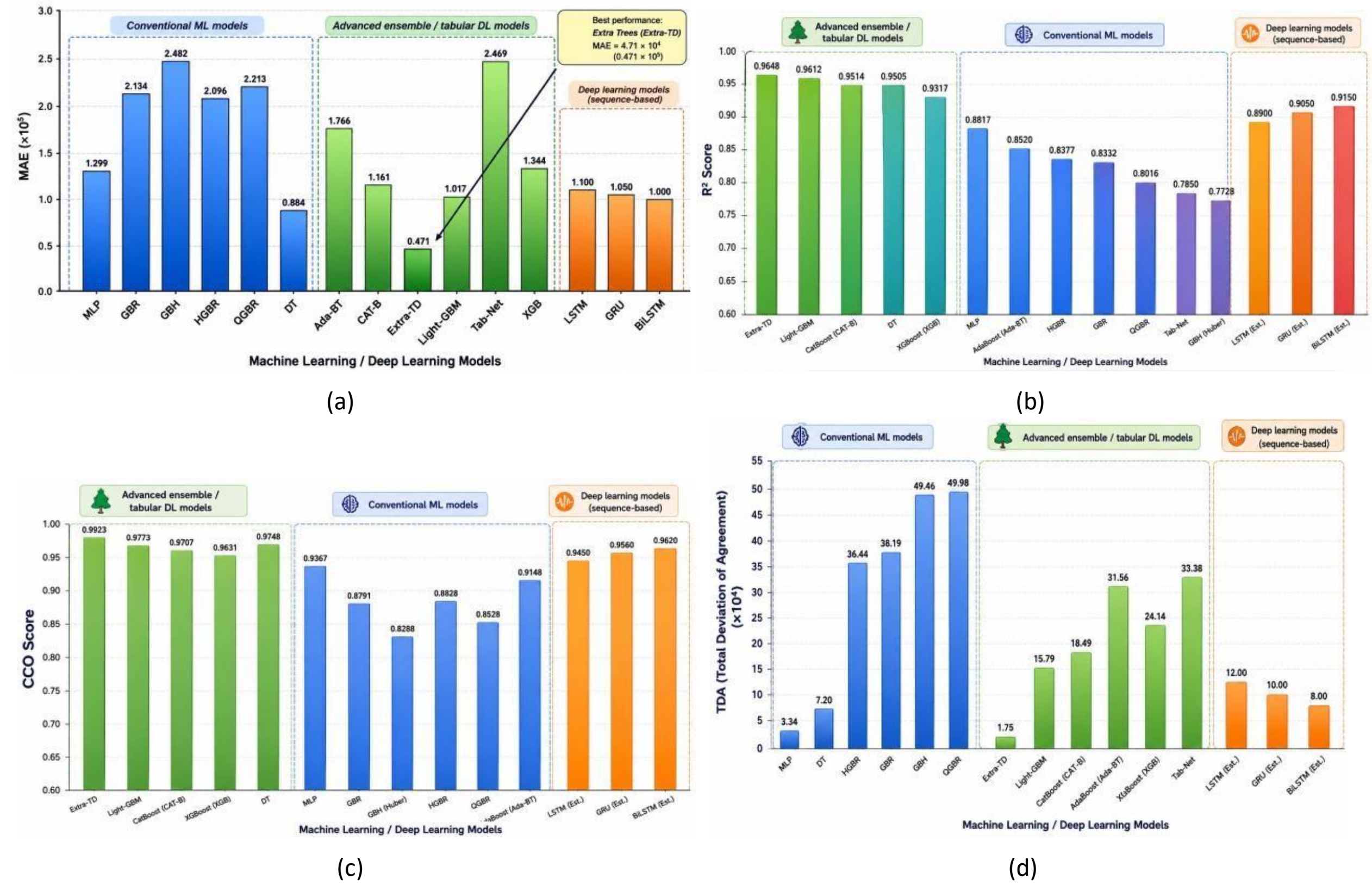

**Figure 4: Comparative performance of 15 machine-learning and deep-learning models for predicting the total power output of 16-WEC layouts at the Perth wave-energy site, evaluated using MAE, $R^2$, CCO, and TDA. Lower MAE and TDA indicate better performance, whereas higher $R^2$ and CCO indicate greater predictive accuracy and agreement.**

Importantly, the results also demonstrate that the model with the highest correlation is not necessarily the most accurate or reliable predictor. XGBoost–BiLSTM achieved the highest mean correlation , but its MAE, EVS, and run-torun variability were less favorable than those of RF–BiLSTM. Similar differences were observed across the other evaluation criteria. Consequently, model selection based on a single metric can lead to incomplete or potentially misleading conclusions. The joint assessment of absolute error, correlation, explained variance, logarithmic error, and prediction stability provides a more reliable basis for evaluating offshore renewable-energy forecasting models.

Overall, the findings indicate that there is no universally superior AI architecture for offshore renewable-energy applications. Instead, the model should be matched to the statistical and physical structure of the problem. Randomized and boosted tree ensembles provide a particularly strong choice for structured WEC surrogate modeling; graph-based networks become advantageous when explicit spatial interactions between energy devices are important; and hybrid ensemble–recurrent architectures are highly competitive when nonlinear tabular relationships coexist with temporal dependencies. This observation provides a practical basis for moving beyond model selection based solely on algorithm popularity or architectural complexity.

Several opportunities remain for further development. Future studies should evaluate the identified models across additional offshore wind and wave sites, longer forecasting horizons, changing environmental regimes, and independent out-of-distribution datasets to establish their robustness and transferability. Uncertainty-aware and probabilistic

(a) (b)

(c)

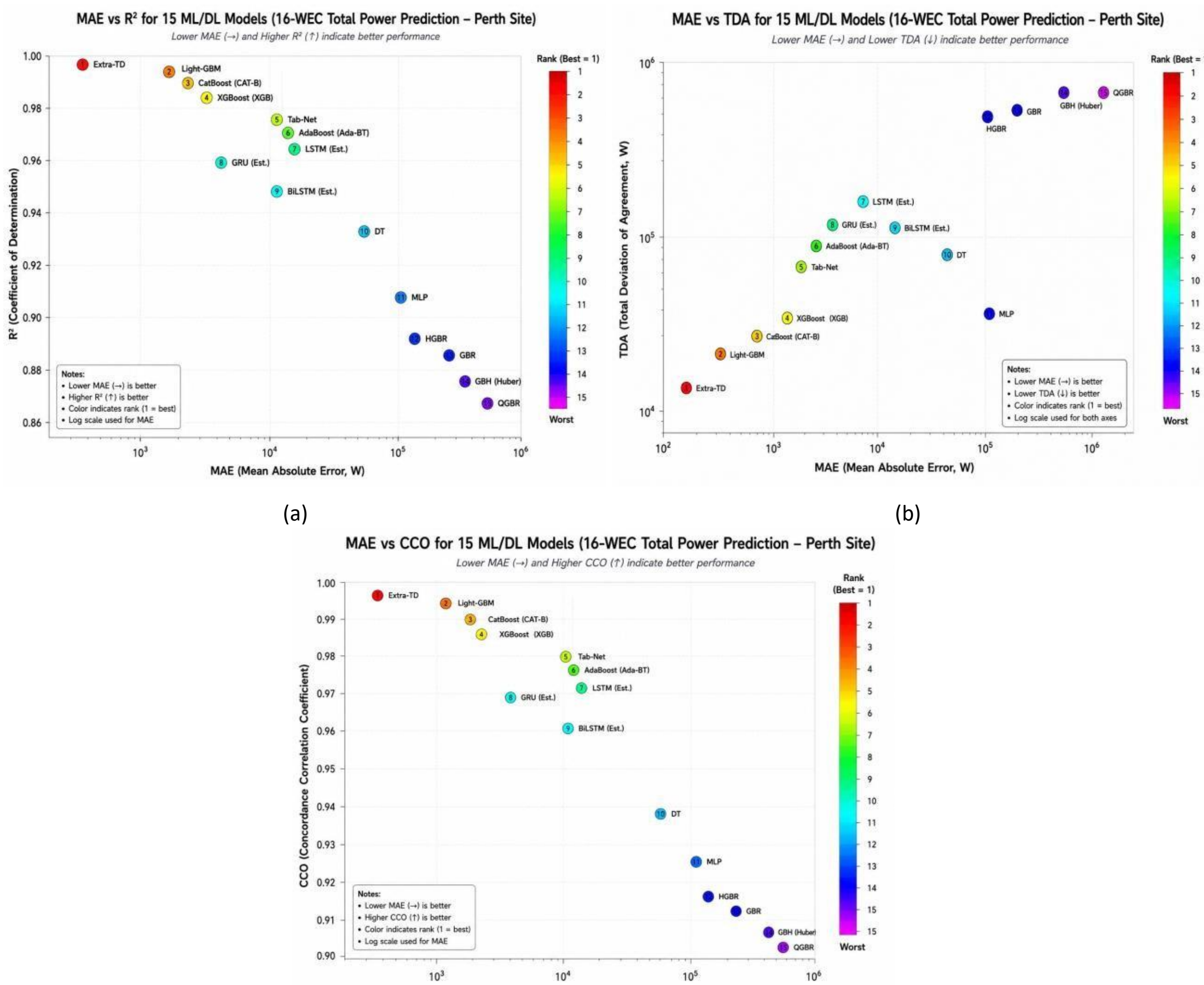

**Figure 5: wo-dimensional performance comparison of 15 ML/DL models for 16-WEC total power prediction at the Perth site based on MAE–$R^2$, MAE–CCO, and MAE–TDA relationships. Marker colors indicate the relative model ranking, with optimal performance characterized by lower MAE and TDA and higher $R^2$ and CCO.**

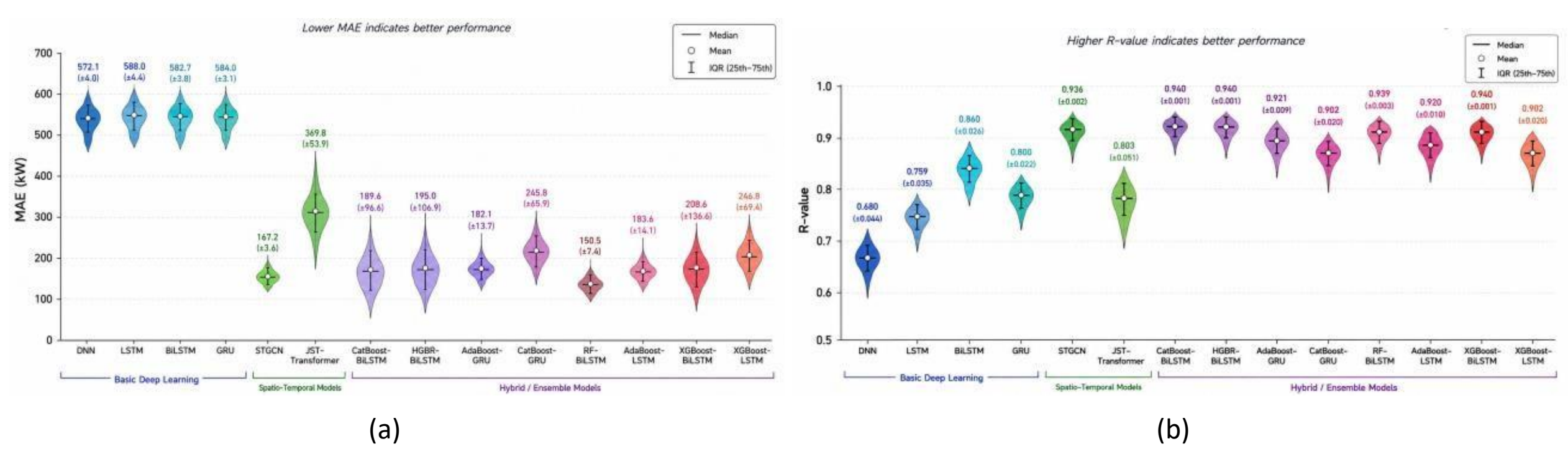

(a) (b)

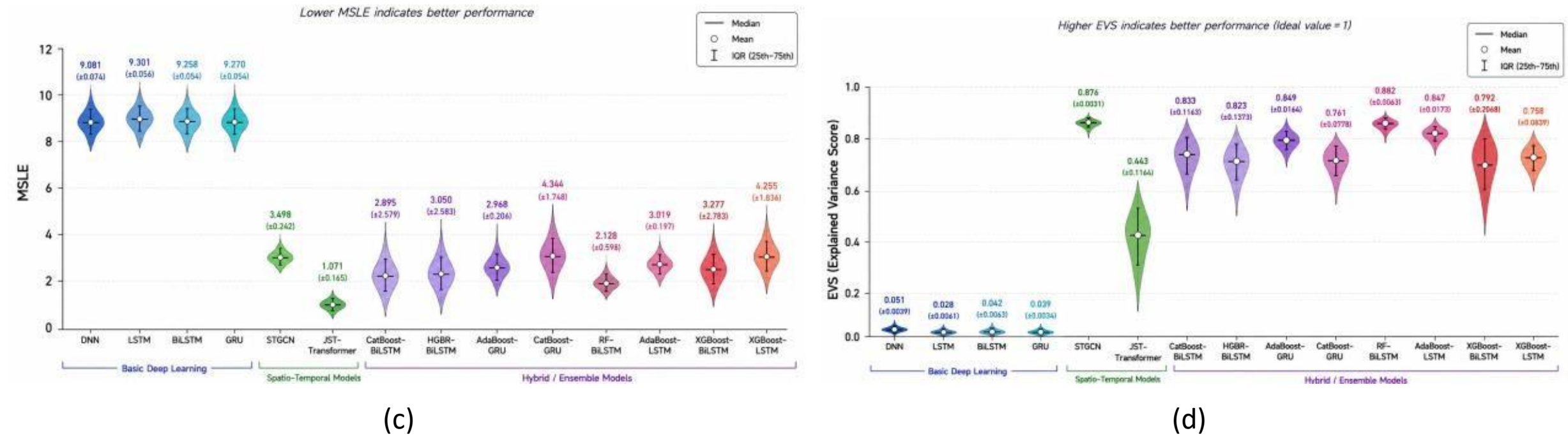

(c) (d)

**Figure 6: Violin-plot comparison of 14 forecasting models for 10-minute-ahead wind power prediction using MAE, $R$-value, MSLE, and EVS, illustrating differences in predictive accuracy, correlation, logarithmic error, explained variance, and performance stability across repeated runs.**

forecasting should also be incorporated to quantify prediction confidence under extreme or sparsely represented operating conditions. From a methodological perspective, physics-informed learning, dynamic graph construction, multi-modal learning, and adaptive hybrid architectures provide promising mechanisms for integrating physical knowledge with data-driven representations. More recent Transformer and time-series foundation models, as well as large language model (LLM)-assisted frameworks, could further support cross-site knowledge transfer, automated feature and model configuration, and interpretable decision support. Finally, integrating high-accuracy surrogate and forecasting models directly within optimization loops represents an important next step toward computationally efficient, adaptive, and reliable design and operation of large-scale offshore renewable energy systems.

## Acknowledgments

Generative AI (GPT 4.0) tools were used solely to assist with English language editing, including grammar, clarity, and readability. All AI-assisted revisions were carefully reviewed and verified by the authors to ensure technical accuracy and preserve the intended scientific meaning.

# Supplementary Material

**Table S1: Comparative analysis of hybrid and advanced evolutionary optimisation methods for wave energy converter (WEC) farm layout optimisation.**

| Ref. | Year | Study | Optimisation method | Decision variables | Farm size | WEC / hydrodynamic model | Objective / metric | Main benefits | Limitations / weak points |
|---|---|---|---|---|---|---|---|---|---|
| [51] | 2018 | Neshat et al., comparative optimisation of wave energy converter placements using multiple metaheuristic methods | (1 + 1) EA, CMA-ES, Differential Evolution (DE), Particle Swarm Optimisation (PSO), Grey Wolf Optimiser (GWO), local search and other metaheuristics | Continuous WEC $x$–$y$ coordinates subject to minimum separation and farmboundary constraints | 4 and 16 WECs | Point-absorber WEC array with hydrodynamic interaction modelling under Australian wave scenarios | Maximise total absorbed power and array interaction performance | Provides an extensive comparative benchmark of evolutionary and swarm-based optimisation approaches for the same WEClayout problem, highlighting differences in convergence behaviour, robustness and evaluation efficiency. | Conventional population-based methods require many expensive hydrodynamic evaluations; scalability deteriorates considerably as the number of WECs and corresponding decisionspace dimensionality increase. |
| [52] | 2019 | Neshat et al., *Adaptive Neuro-SurrogateBased Optimisation Method for Wave Energy Converter Placement Optimisation* | Adaptive metaheuristic optimisation + recurrent neuralnetwork surrogate + greedy local search / back- tracking | Continuous WEC $x$– $y$ locations | 4 and 16 WECs | Point-absorber WEC farm with computational hydrodynamic model; surrogate model used to approximate expensive power evaluations | Maximise total farm power while reducing the number of expensive hydrodynamic evaluations | Introduces an AI-assisted evolutionary optimisation framework in which a neural surrogate approximates costly WEC-array evaluations; adaptive search and local refinement improve computational efficiency. | Optimisation performance depends strongly on surrogate accuracy and training-data representativeness; inaccurate surrogate predictions may guide the optimiser toward misleading regions of the search space. |
| [53] | 2019–2020 | Neshat et al., hybrid local-search frame- work for position optimisation of wave energy converters | Hybrid stochastic / symmetric local search + Nelder– Mead search + knowledgebased sequential optimisation | Continuous WEC $x$– $y$ coordinates with sequential placement and local refinement | 4 and 16 WECs | Point-absorber WECarray model evaluated under simplified and realistic Australian wave climates | Maximise total absorbed wave power using a limited computational budget | Highly evaluation-efficient compared with conventional populationbased metaheuristics; combines exploitation-oriented local search with domain knowledge and sequential placement to rapidly identify high-quality arrangements. | Sequential and greedy placement may introduce dependency on early positioning decisions and can become trapped in local optima; backtracking or global exploration is needed to correct previously selected locations. |
| [54] | 2020 | Neshat et al., *A hybrid cooperative coevolution algorithm framework for optimising power takeoff and placements of wave energy converters* | Hybrid Cooperative Coevolution (CC) + evolutionary optimisation + local search | WEC $x$–$y$ locations together with individual PTO damping / control parameters | 4, 16 and larger benchmark WEC farms | Point-absorber array with hydrodynamic interaction and individual PTO mod- elling under realistic wave climates | Jointly maximise wave-farm power through simultaneous layout and PTO optimisation | Decomposes the high-dimensional joint layout–control problem into interacting subcomponents, allowing spatial and PTO variables to be optimised more efficiently than with monolithic evolutionary optimisation. | Performance depends on variable decomposition and interaction grouping; cooperative optimisation increases algorithmic complexity and may still require substantial computational resources for large farms. |
| [8] | 2022 | Neshat et al., *Layout optimisation of offshore wave energy converters using a novel multiswarm cooperative algorithm with backtracking strategy* | Multi-Swarm Cooperative Co-evolution (MSCC) combining multiple populationbased optimisers, surrogate mod- elling and backtracking | Continuous WEC $x$–$y$ locations subject to farm-area and minimum-separation constraints | 4 9 and WECs | Point-absorber WECarray hydrodynamic model evaluated under six realistic Australian wave-energy sites | Maximise total absorbed power and improve convergence under limited evaluation budgets | Combines complementary exploration and exploitation behaviours of multiple optimisers; cooperative search, surrogate assistance and backtracking improve convergence and allow poorly positioned WECs to be reconsidered during optimisation. | The framework contains several interacting search components and control parameters, increasing implementation and tuning complexity; experimental farm sizes remain relatively small compared with future commercialscale arrays. |